# Probabilistic Warnings in National Security Crises: Pearl Harbor Revisited

David M. Blum[*]     M. Elisabeth Paté-Cornell[#]

## ABSTRACT

Imagine a situation where a group of adversaries is preparing an attack on the United States or U.S. interests. An intelligence analyst has observed some signals, but the situation is rapidly changing. The analyst faces the decision to alert a principal decision maker that an attack is imminent, or to wait until more is known about the situation. This warning decision is based on the analyst's observation and evaluation of signals, independent or correlated, and on her updating of the prior probabilities of possible scenarios and their outcomes. The warning decision also depends on the analyst's assessment of the crisis' dynamics and perception of the preferences of the principal decision maker, as well as the lead time needed for an appropriate response. This article presents a model to support this analyst's dynamic warning decision. As with most problems involving warning, the key is to manage the tradeoffs between false positives and false negatives given the probabilities and the consequences of intelligence failures of both types. The model is illustrated by revisiting the case of the attack on Pearl Harbor in December 1941. It shows that the radio silence of the Japanese fleet carried considerable information (Sir Arthur Conan Doyle's "dog in the night" problem), which was misinterpreted at the time. Even though the probabilities of different attacks were relatively low, their consequences were such that the Bayesian dynamic reasoning described here may have provided valuable information to key decision makers.

[*] Applied Economics Program, Johns Hopkins University, Washington, DC 20036, dblum7@jhu.edu

[#] Department of Management Science and Engineering, Stanford University, Stanford, California 94305, mep@stanford.edu

# 1. PROJECTING AND WARNING

The problems of anticipating and predicting the future moves of an adversary date to the earliest recorded military history. Mathematical probabilistic analysis of some of these problems began in the 19th century with the work of Antoine Cournot and proceeded further before and during World War II, leading to an extensive theory of games. As early as 1942, the British secret services at Bletchley Park used their ability to decrypt German Enigma communications, newly developed Bayesian methods, and the first computers at their disposal to protect transatlantic convoys from U-Boat "wolf packs" in the Northern Atlantic. At the height of the Cold War, Schelling (1960), among others, was advocating the use of game theory to conceive and implement deterrence strategies. The world is now facing a number of problems, some related to rivalries among nationstates, some to terrorism and crime, for which information can be gathered and analyzed to warn of future developments and guide immediate decisions.

The purpose of this article is to present a dynamic, Bayesian approach to early warning analysis in that intelligence context. We propose a formal model to support a probabilistic information system (Edwards et al. 1968) and guide intelligence analysts in deciding when to issue warnings of imminent events, called tactical warning in the military context (Grabo 1994), to principal national security decision makers. The goal is to give the principals the lead time that their contingency plans require and minimize the chances of false alerts, accounting for both temporal and spatial evolution of a threat. The model we propose represents an advance in the state of the art of intelligence warning analysis because it involves both dynamic systems and decision analysis. It includes:

1. Incorporation of geographic specificity of threats to gain localized warning;
2. Modeling crisis dynamics to assess lead times;
3. Representation of probabilistic dependencies among variables and signals;
4. The analyst's warning decision of when and how to alert a principal.

A model that addresses the first two features can be formulated using recursive Bayesian estimation (West and Harrison 1989). The treatment of probabilistic dependencies and the time-dependent decision, however, are sources of mathematical complexity. Whereas the concepts are relatively straightforward, a real-world implementation is likely to require computer software to aid the intelligence analyst. Our focus is on the individual analyst as the user of this model to make warning decisions. Reconciling views of multiple analysts is thus beyond our scope. For the same reason, and because our model is intended to be prescriptive rather than descriptive, principal-agent preference misalignment is also beyond our scope. We assume that the principal and the analyst agree on a set of preferences. Both issues are worthy of further studies to better understand the causes of some intelligence failures.

The timing of any warning must balance the risk of a false alert with that of a missed one, accounting for the lead time that the response plans assume (Paté-Cornell 1986). In our model, how crises evolve over time, what we call crisis dynamics, holds the key to lead-time inference. We first formulate a dynamic model of warning decisions from the point of view of an intelligence analyst. We then illustrate it by an analysis of the immediate lead up to the historic Japanese surprise attack on Pearl Harbor on December 7, 1941, concluding with a discussion of some issues surrounding real-world implementations of this model.

## 2. DYNAMIC WARNING MODEL

The optimization of an imperfect warning system relies on a tradeoff between the probabilities of a false positive (false alert) and of a false negative that fails to convey an alert or to meet a lead time requirement (Peterson et al. 1954, Swets and Birdsall 1956). Facing an evolving crisis, an analyst may employ a thought process whose steps involve:

1. Observing new intelligence signals relevant to the current unknown threat state and updating the probabilities of different possible states;
2. Applying beliefs about crisis dynamics to the probability of the current threat state to extrapolate a probability distribution of the threat state in future time periods;
3. Utilizing an understanding of the principal's preferences (i.e., costs associated with outcomes, risk attitude, and time discounting) and of the extent to which future intelligence could change the probabilities of different threat states to decide whether to issue an alert of a given type (e.g., severity level) right away or wait for additional signals; then repeating the process in the next time period.

The model that supports this process includes four submodels (see Table 1 and Figure 1).

**Table 1 Four Submodels in Warning Analytic Framework**

| Submodel | Purpose |
|---|---|
| Crisis definition submodel (CDM) | Construct a prior joint probability distribution of fundamental, static variables that affect the manner in which a crisis evolves (e.g., adversary intentions and capabilities). |
| Inference submodel (IM) | Infer the probabilities of the different possible current threat states from available information, and update probability distributions over static variables |
| Basic warning submodel (BWM) | Derive a probability distribution of crisis timing and type or severity in future time periods. |
| Disutility submodel (DM) | Decide which type of alert to issue now or whether to wait, given the preferences of the principal and the anticipated value of information that can be gathered in the future. |

The crisis definition submodel (CDM) contains the static elements relevant to a crisis situation. If some dynamic variables (e.g., a potential change of leadership and preferences, or adversary's military capabilities) need to be included in order to streamline the basic warning submodel (BWM), or if assumed static variables change while the model is running, the CDM can be reinitialized accordingly when those variables change. The BWM contains a Markov process that characterizes the dynamic variables needed to project ahead the system's state (e.g., the state of an adversary's attack planning). It is linked to the CDM by a static joint random variable representing those fundamental uncertainties that influence the manner in which a crisis may evolve in the future, for example, an adversary's capabilities and intentions. We use the concept of conditional independence, and more precisely "d-separation" as defined in the study of Bayesian networks to describe an algorithm in which the influence of random variable X on variable Y flows exclusively through variable Z (Koller and Friedman 2009). As such, we refer to this joint variable as the "D-separating joint variable," or simply as "D," because of its structural position between the CDM and all that comes after it (Geiger et al. 1990).

# Figure 1 Relationships Among the Four Submodels

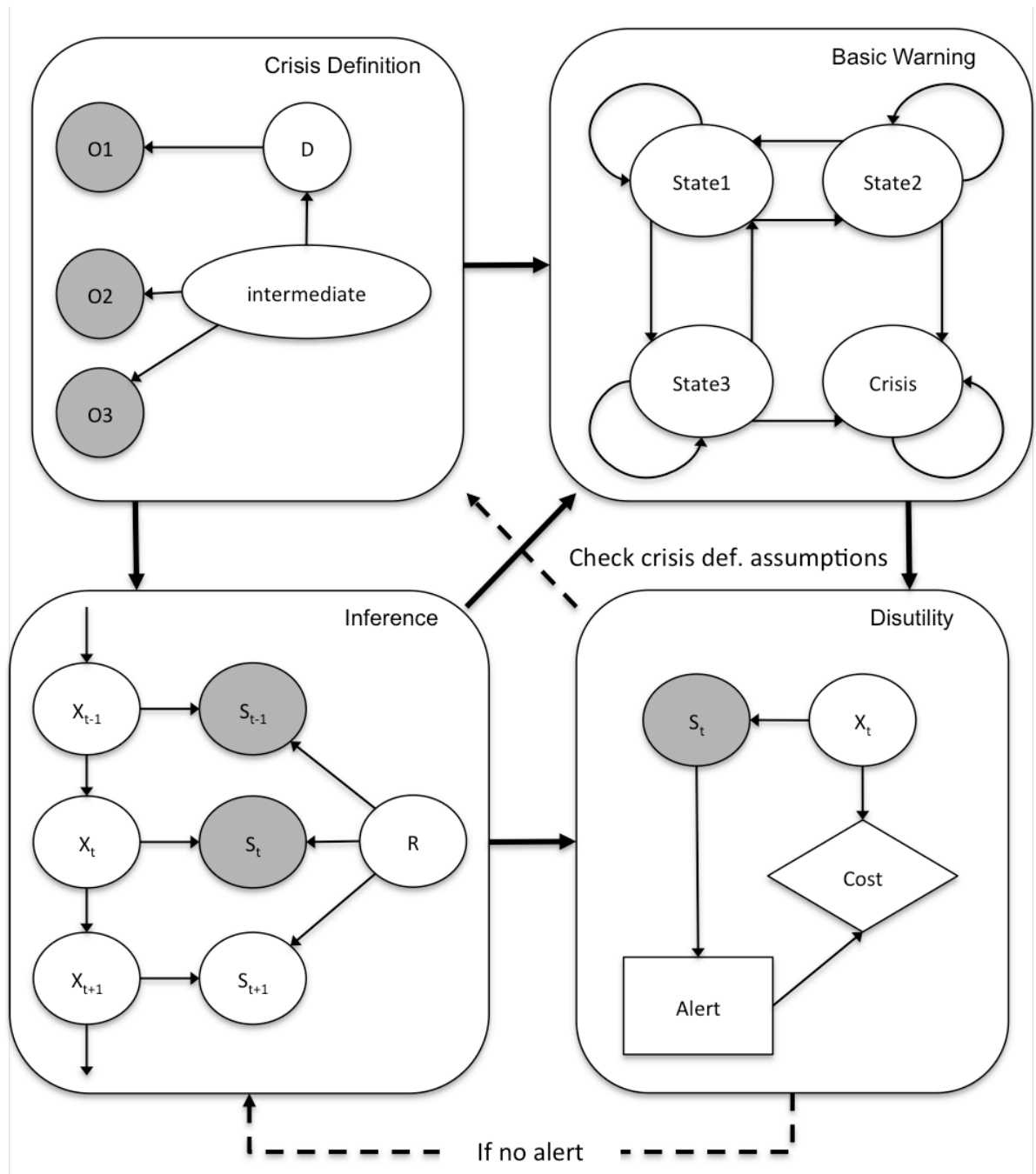

The inference submodel (IM) is a modified hidden Markov model (Baum and Petrie 1966, Baum and Eagon 1967) based on the same underlying dynamic process as the BWM. It yields a probability distribution of the present state of the dynamic system given imperfect signals and performs the function of incorporating signals relevant to the evolution of a crisis into machine memory (Paté-Cornell and Fischbeck 1995). The disutility submodel (DM) defines the recurring decisions of the analyst in successive time periods of whether to issue one of several predefined alerts to a principal decision or to allow the crisis to evolve further in order to gather more information. We assume that alert levels correspond to potential responses that have already been determined and incorporated into deliberate plans, which may reflect means of attack, severity, or any other salient feature that will drive the response. Note that the BWM state space will have to be defined in such a way that these features correspond to one or more states. The lead time that response plans require for implementation and the cost of missed alert are exogenously specified. One key variable of this model is the "disutility" (negative expected consequence) attributed by the principal to adverse outcomes and costs.

The first three submodels yield a probability distribution of future crisis occurrences and locations (if relevant). The fourth submodel uses that distribution as an input to a decision problem, assuming that the analyst is given the responsibility to issue various types of alerts. Such delegation is common, but not universal, in the world of intelligence (Belden 1977). The problem is solved in accordance with the principles of decision analysis (Howard and Matheson 2005). This fourth submodel leverages concepts developed by Peterson, Birdsall, and others in the 1950s, specifically the balancing of costs associated with true positive and false positive detections of signals embedded in noise, together with the costs associated with true negative and false negative "nondetections." However, we do not implement their optimal signal receiver because their model assumed prior knowledge of the probability densities of samples associated with noise and with signal added to noise, which were fixed across the time interval of analysis. We believe that a dynamic Bayesian model is better suited to the warning intelligence application because it allows analysts to assess conditional probabilities associated with each signal individually *as it is observed*, and to use it to update their prior belief of the crisis' state in accordance with Bayes law. Central Intelligence Agency methodologists have cited this as a helpful feature of Bayesian methods in intelligence analysis (Zlotnick 1972, Heuer 1981). The analyst need not assess the sample probability densities over fixed periods of time, which would be impractical. Table 2 summarizes the features and variables of all four submodels.

## 2.1. Formulation of the Crisis Definition Submodel

The CDM is represented by an influence diagram used to infer Prob $D|O$, where $O$ is the set of observations relevant to $D$, that is the set of jointly distributed static random variables (hereafter, joint variable) relevant to a crisis. A given realization of $D$ implies a particular set of transition probabilities for the underlying dynamic process of the crisis that is unfolding. Typically, the joint variable $D$ represents the intentions and capabilities of an adversary during the time in which a crisis is evolving. Several observations comprising the set $O$ may be relevant to these static fundamentals. If necessary, intermediate random variables can be defined to facilitate inferring Prob $D|O$, a process called "knowledge mapping" by Howard (1989). For example, if the crisis that an analyst is tasked to monitor involves the progress of a foreign nation's nuclear program, $D$ might refer jointly to (1) that nation's nuclear objectives and (2) its scientific industrial base. An observation relevant to these static fundamentals might include that a decades-old war between the nation and one of its neighbors was terminated by armistice but never codified by treaty. The existence of threats to the nation's territorial integrity might serve as a useful intermediate variable to assist in incorporating this observation. Although a change in a variable that was assumed to be static will necessitate reinitialization of the model, this should be a relatively infrequent occurrence as long as the model is used for tactical warning rather than long-term, strategic warning. The crisis definition node of Figure 1 illustrates a generic CDM used to assess the joint variable $D$ using an intermediate variable and three relevant observations.

**Table 2 Summary of Expressions and Features, Including Inputs and Outputs**

| Notation | Indices | Input or feature of | Output of | Description |
|---|---|---|---|---|
| D *joint random variable* | -- | BWM IM DM | CDM | The set of jointly distributed static random variables relevant to state $X_t$ (called here the *D-separating joint variable*). We use it to represent (approximately) static unknowns on which crisis evolution depends, such as the adversary's intentions and capabilities. Realizations of D are indexed by j. It is assumed that D contains all the information needed to derive transition probabilities among crisis states. |
| {O} *set* | -- | CDM | -- | Set of all observed events relevant to D. |
| **X** *set of states* | Elements indexed by k | BWM IM | -- | *State space* characterizing all possible states of the dynamic variable driving an evolving crisis. Those states k which are trapping states comprise set {k}. |
| **P** *vector of transition matrices* | Elements of **P** indexed by j (for each realization of D) | BWM IM DM | -- | A vector of matrices, each containing *transition probabilities* among states in **X** in a unit of time given a realization j of D. An element of **P** is a matrix corresponding to j and denoted $P_j$. Every matrix $P_j$ includes one or more trapping states k∈{k}. |
| $X_t$ *random variable* | t (time period) | BWM IM | -- | The dynamic random variable representing the state of an evolving crisis in space **X** at a particular time t. |
| $H_{k,t}$ *random variable* | {k} (trapping states), t (time period) | DM | BWM | A random variable representing the *future time* of first passage to state k of an unfolding crisis, beginning at time t, i.e. the present time. |
| T *scalar* | -- | DM | -- | An arbitrary time horizon in the future. |
| $S_t$ *random variable* | t (time period) | IM | -- | Noisy intelligence signal observed in time period t. It depends on the current state $X_t$ and on the credibility R of available intelligence sources (see below). |
| R *random vector* | -- | IM | -- | A vector of random variables whose elements correspond to the veracity (0 or 1) or measurement error (continuous values between 0 and 1) of a unique source of one or more signals $S_0$,..., $S_t$ (source is defined broadly). |
| **p₀** *vector* | Elements indexed by k (state of **X**) | IM | -- | Vector of initial probabilities of the state of $X_t$ at the time the inference submodel (IM) is initialized. |
| **π**$_t$ *vector* | t (time period); elements indexed by k (state of **X**) | BWM DM | IM | Vector of posterior probabilities of state $X_t$ at time t incorporating all signals $S_0$,..., $S_t$ gathered up to time t. |
| **V** *matrix* | Elements indexed by j (realization of D), k (state of **X**) | DM | -- | Matrix containing dollar-equivalent or multi-attribute *failure costs*. Cost $v_{j\{k\}}$ is incurred when the analyst fails to give a warning inside of the minimum lead time before the crisis hits a trapping state k ∈ {k}. For all transient states (i.e., states not in set {k}), cost $v_{jk}$ = 0. |
| **q** *vector* | Elements indexed by i (type of alert) | DM | -- | Vector of *alert costs*, which are fixed cost associated with each type of alert i, whether true or false (e.g. value of lost training). Costs $q_i$ use the same attributes as $v_{jk}$. There may be up to as many types of alerts as responses considered. |
| **L** *vector of matrices* | Elements indexed by i (type of alert), j (states of D), k (state of **X**) | DM | -- | A vector of matrices containing the *minimum necessary lead times for response options*. Lead times $l_{ij\{k\}}$ represent the number of time periods that the principal's response assumes will be available between an alert and a crisis entering one of its trapping states in order to achieve success. If $l_{ijk} = 0$, then no warning lead time is necessary to avoid incurring a failure cost. |
| α *scalar* | -- | DM | -- | Principal's *discount rate*. The present value at time t of a cost v incurred in time t+n is $v/(1+\alpha)^n$. |
| a*(**p₀**,{S}) *function* | -- | -- | DM | The analyst's optimal alert decision (time and type) as a function of initial probabilities of crisis states and interpretation of signals observed. |

*Note.* BWM: basic warning submodel; CDM: crisis definition submodel; IM: inference submodel; DM: disutility submodel.

## 2.2. Formulation of the Basic Warning Submodel

The analyst defines a state space X that fully characterizes the crisis situation as it unfolds, and thus the underlying dynamic process (such as the status of an adversary's attack plan). The state space **X** must be comprised of the same set of states given any possible realization of the joint variable D. Consider again a dynamic process characterizing a foreign nation's evolving nuclear program, which includes a joint variable D representing (1) its nuclear objectives and (2) its scientific industrial base. A realization of that joint variable might be (1a) achievement of energy independence, together with (2a) lacking a scientific industrial base sufficient to reprocess plutonium to make a weapon. A different realization of D might be (1b) ensuring territorial integrity and (2b) an industrial base capable of building a plutonium reprocessing plant. Turning now to the state space **X**, one state might involve the presence of a light-water reactor along with a small centrifuge cascade optimized to produce fuel for this reactor. An analyst might treat this state as a trapping state given the realization of D (1a, 2a) but transient given realization (1b, 2b). Given (1a, 2a), the analyst does not consider any further evolution of the system once it reaches this state, but does consider the possibility of further crisis developments given (1b, 2b), for instance, a transition to a state in which the nation possesses a nuclear weapon. We require **X** to be defined so that given the state $X_t$ of the crisis at time t and the intelligence sources R, it satisfies the Markov property with a single step of memory in transitions among states.[1]

If the crisis enters a trapping state belonging to set k at some future time, implying that no further evolution is considered, it must be for the first time. We can then compute the probability distribution over the future "first passage time" from an uncertain state at time t to any of the trapping states in k given the realization j of the joint variable D (representing the set of fundamental static variables affecting how a crisis unfolds).

Consider:

- $\boldsymbol{\pi}_t$: the probabilities, as assessed by the analyst, of being in state $X_t = k$ (a particular state) at a particular time t. It is an output of the IM and is described in the next section.
- $P_j$: a matrix of state transition probabilities among states comprising **X** given realization j of joint variable D.
- T: an arbitrary distant time horizon beyond time t.
- τ: a *running variable for future time* at which the evolving crisis enters state k.
- The subscript $_{k:0}$ appended to a vector, which we use to indicate that the value 0 is inserted in place of the k$^{th}$ element of vector $(\boldsymbol{\pi}_t P_j^{\tau-1})$.
- A second subscript $_k$ appended to a vector, which we use to represent the scalar value of the k$^{th}$ element of vector $((\boldsymbol{\pi}_t P_j^{\tau-1})_{k:0} \boldsymbol{P}_j)$.

The probability of first passage time for any state in {k} given D=j, in each time period τ between t (present time) and T (time horizon), is:

$$Prob\{H_{\{k\},t}(\tau) | D = j\} = \sum_{k \in \{k\}} ((\boldsymbol{\pi}_t P_j^{\tau-1})_{k:0} \boldsymbol{P}_j)_k \qquad (1)$$

The sum over k ∈{k} represents adding up the probabilities that the process transitions to each trapping state. The results of Equation (1) are inputs to the DM.

The nuclear example mentioned above illustrates how a state space might be defined reflecting a crisis whose dynamics are not tied to geography. Crises in which the adversary's location changes and reflects the state of the crisis are an important class of tactical warning problems. The illustration presented in Section 3 (the lead up to Pearl Harbor) is a crisis of this type. For this class of problems, an appropriate state space is the location of the entity (here, the Japanese carrier strike force) given some static variable (here, the strike force's next target and the speed at which it is traveling to reach that target). For a multidimensional geographic space, such as an entity's movement by sea or air, one might discretize the space into a lattice of triangles called rasters (or into volumes) and define a crisis state as the raster inside which the entity is located at a given time. Transition probabilities are nonzero only for rasters inside a range determined by the entity's maximum speed. When the entity's motion is constrained to one dimension, for instance by movement along known roads, one might define a network of permissible travel routes and a state as a node where the entity may be located at a given time. All transition probabilities to nonadjacent nodes are equal to zero.[2]

## 2.3. Formulation of the Inference Submodel

A modified hidden Markov model is constructed from the random variables $X_t$ and $S_t$ (representing observed signals relevant to $X_t$), and the random vector R (representing signal source credibility), as shown in the inference node of Figure 1, given the initial probability distribution **p₀** over $X_0$. The process is "hidden" because the analyst does not observe the state but does observe noisy signals that are relevant to the state, following the Markov property as in the BWM.

A random vector R contains m elements reflecting the credibility of m unique intelligence sources, each common to one or more signals. The realizations of R might be binary in cases where a source's credibility is measured solely by whether or not it reports truthfully (e.g., a specific spy). Alternatively, the credibility levels might take on continuous values between 0 and 1 in cases where a source is graded based on its measurement error, which may be appropriate for technical sensors. In subsequent calculations, we assume discrete realizations of elements of R, which implies discrete measurement errors of arbitrary granularity, but the model can be easily extended to the continuous case.

Sources whose credibility is represented by a single element of the vector R are assumed to be conditionally independent of one another given the crisis state. Therefore, the quality of the signal or message from each source can be assessed in isolation prior to any signal observations, then updated using Bayes rule as new signals are observed. Dependent sources are treated as a single source whose corresponding element in R has an appropriate number of realizations to reflect the credibility of the "subsources."[3] If all of the m elements of R are binary and contain no dependent subsources, the vector R has $2^m$ possible realizations. If a source or set of subsources is represented as a distinct element of R, it is assumed that this source or set does not overlap with any others represented in other elements. All dependencies are captured within the credibility levels of each element. If the source of a signal is not identified, or might be one of several sources, this may prove impossible without assuming that the unidentified source is a unique source (possibly a dubious assumption).

This formulation of R as the sources' credibility allows assessment of the quality of a signal given the state of a crisis where signal dependence is due to common sources. If, however, signal dependence is thought to have a more nuanced meaning, the relevant elements of R may require a different specification based on the analyst's beliefs. For example, reports derived from radio intercepts of conversations might require a specification of R that reflects both the likelihood that the conversing individuals are specific known actors (e.g., scientists who work at a centrifuge facility) as well as the likelihood that those known actors have knowledge of ground truth (e.g., the size and arrangement of the centrifuge cascade).

For a classic hidden Markov model, the usual approach to incorporating signals $S_0$, ... , $S_t$ into one's probability of the present state $X_t$, is referred to as the forward algorithm (Baum and Eagon 1967). That algorithm involves applying Bayes' rule, using the signal observed at time 0 to update the prior belief of the dynamic process, then applying transition probabilities to obtain a new probability for the state of the process at time 1, and iterating. This algorithm is applied here with minor modifications to solve for Prob $\{X_t \mid D = j, \mathbf{p_0}, S_0, ... , S_t\}$, where we have added, through R, a variable representing dependence among common, imperfect sources.

Let Prob{R} be the probability distribution of the random vector R (credibility of sources). We then substitute $\sum_R (\text{Prob}\{R\} \text{Prob}\{S_t \mid R, X_t\})$ in place of Prob$\{S_t \mid X_t\}$ in the forward algorithm and also update Prob$\{R \mid S_t\}$ using Bayes rule after observing each successive signal. This is possible because future signals still do not contain any information, as demonstrated by the application of Shachter's Bayes Ball algorithm to the network shown inside the inference node of Figure 1 (Shachter 1998). The analyst's belief, Prob{R}, must contain all necessary information, and thus includes the validity of all signals. Using the modified forward algorithm, we update the probabilities of the crisis states given the signals to date. We then multiply these terms and sum over the static realizations of D to obtain the distribution of the crisis states at time t, and we organize the distribution into a vector of posterior probabilities assessed at time t:

$$\boldsymbol{\pi_t} = \sum_D (Prob\{X_t|\boldsymbol{p_0}, D = j, S_0, \ldots, S_t\} Prob\{D|S_0, \ldots, S_t\}) \quad (2)$$

Recall that the node R is updated whenever a new signal $S_t$ is observed. For example, new intelligence from a source allows an analyst to reevaluate the validity of past reporting from the same source. To account for this, whenever a new signal is observed from a source with a reporting history, one must iterate the Forward Algorithm between the first time the source reported and time t using the updated probabilities for R. Dependencies among signals thus complicate significantly the analysis of a warning model.

### 2.4. Formulation of the Disutility Submodel (Costs and Risk Attitude)

Besides the correspondence of warnings to planned responses, we make here several assumptions. First, the principal must be a rational decision maker in the classical (von Neumann) sense. Second, the analyst knows and has adopted the principal's costs, meaning willingness to pay for different outcomes resulting from true alerts, false alerts, and missed alerts, which include alerts given after the minimum necessary lead time, as well as the principal's risk attitude and time preference. We make this assumption because this warning model is intended to be normative, and analysts are not supposed to express their own preferences. Third, the *disutility* function U, common to the analyst and the principal, reflects a constant absolute risk aversion, which implies a linear or exponential disutility function. The time preference common to principal and analyst is represented by a constant discount factor α. Costs might be assessed in monetary-equivalent terms if it is convenient, or they might involve multiple attributes. In that case, we use a multi-attribute value function. The principal's indifference curves must be specified, to form a value function, and his or her risk attitude must be expressed over that value function (Matheson and Abbas 2005). Fourth, by convention, we assess the cost of the expected outcome of a true alert to be 0 plus the cost **q** of issuing the alert (i.e., the cost of response), and we make the simplifying assumption that **q** will be the same whether the alert is true or false.

This fourth assumption, although perhaps unconventional, has little effect on the model's results because we have already assumed the existence of a crisis trapping state, meaning that crises will occur eventually with certainty. An analyst is incentivized to put off telling the principal, "The crisis is imminent, and the time to act is now or never," because time discounting reflects a general desire of principals to delay commitment to action until it is absolutely essential. No alert is ever truly false in this model, only premature! This assumption implies that a correct nonalert has a zero cost. It also implies that fixed alert costs must be greater than 0. A willingness to give away some value beneath a specified threshold (in football, a "prevent defense") is not possible if any alert cost is beneath that threshold. If it were, the analyst's incentive would be to ignore the possibility of a false alert and immediately issue a "free" alert corresponding to the worst crisis state whose alert cost is beneath the threshold.

The analyst's decision problem is formulated as a particular Markov decision process, a minimization of expected disutilities, in which the analyst's actions affect the rewards for each state but not the transition probabilities (Howard 1960). Bellman's implementation of value iteration for Markov decision processes is used to solve the decision problem (Bellman 1953). This approach requires considering outcomes at a future time horizon T, beyond which one is not concerned with the same problem.

Consider the probability $\text{Prob}\{H_{\{k\},t}(\tau)|D=j, \mathbf{p}_0, S_0,\ldots, S_t\}$, assessed at time t, that the crisis first hits any trapping state in set k in time periods in the future given the static fundamental variables (D) relevant to a crisis unfolding and the set of signals observed so far $\{S_0,\ldots, S_t\}$. Extending Equation 1 to incorporate signals observed so far yields:

$$Prob\{H_{\{k\},t}(\tau)|D=j, \mathbf{p}_o, S_0, \ldots, S_t\} = \sum_{k \in \{k\}} \left(\left((\boldsymbol{\pi}_t|\mathbf{p}_o, S_0, \ldots, S_t)\mathbf{P}_j^{\tau-1}\right)_{k:0} \mathbf{P}_j\right)_k \quad (3)$$

Let $CE(a_{ti}|D=j, \mathbf{p}_0, S_0,\ldots, S_t)$ denote the certain equivalent, common to analyst and principal, of the overall cost associated with an alert of type i given at time period t, in which their common disutility function is represented as U (with inverse $U^{-1}$). Here t represents a *particular time* period at which the alert is given and the index τ functions as a running variable for future time at which the evolving crisis enters state k. It is assumed that the decision to give an alert is made at time t after observing all signals by minimizing the certain equivalent of the disutility associated with each alert type if it were issued at time t.

$$CE(a_{ti}|D=j, \boldsymbol{p_o}, S_0, \ldots, S_t) = PV_\alpha^t(q_i) + U^{-1}\left(\sum_{\tau=t}^{t+l_{ij\{k\}}} \left[Prob\{H_{\{k\},t}(\tau)|D=j, \boldsymbol{p_o}, S_0, \ldots, S_t\}U\left(PV_\alpha^\tau(v_{j\{k\}})\right)\right]\right) \quad (4)$$

Equation (4) reflects the fact that the certain equivalent $CE(a_{ti}|D=j, \boldsymbol{p_0}, S_0, \ldots, S_t)$ the fixed cost associated with an alert at time t (i.e., $PV_\alpha^t(q_i)$), and an uncertain cost that depends on whether the alert comes too late for a meaningful response, $PV_\alpha^\tau(v_{j\{k\}})$. Because the cost $v_j$ for entering any transient state not belonging to k is 0, only the trapping states contribute to the uncertain portion of $CE(a_{ti}|D=j, \boldsymbol{p_0}, S_0, \ldots, S_t)$. The summation yields the certain equivalent of the failure costs and reflects the uncertainty regarding whether the crisis enters a trapping state within the minimum lead times for planned responses. The analyst's decision problem of the best alert and best time to issue that alert a can then be written as:

$$a^*(\boldsymbol{p_o}, S_0, \ldots, S_t) = \underset{i,\tau:[t,T]}{\mathrm{argmin}}\left[\sum_D Prob(D|S_0, \ldots, S_t)U\big(CE(a_{\tau i}|D=j, \boldsymbol{p_o}, S_0, \ldots, S_t)\big)\right] \quad (5)$$

Equation (5) states that the analyst must find the type and time of alert that yields the lowest expected disutility to the principal, for all alert types i and at all time periods between the present time t and the time horizon T.

The objective function to be minimized is not convex. Therefore, no attempt is made to perform this minimization analytically. In practice, however, it is simple to implement the optimization numerically by simulation, although the speed of computations depends on the extent of dependencies among signal sources (it runs in linear time with respect to t in the special case where there are no dependencies). If the time τ that minimizes the objective function for alert type i is the present time, an alert of type i should be issued right away to the principal. If the time that minimizes the disutility is beyond the present, it is better to wait, observe the next signal, and repeat the optimization rather than risk a false positive.

## 3. RETROSPECTIVE ILLUSTRATION: THE CRISIS IN THE PACIFIC IN NOVEMBER, 1941

We illustrate this analytic framework by considering the historical warning decision problem faced by analysts in the US Navy's War Plans Division (WPD), beginning ten days before Japan attacked Pearl Harbor on December 7, 1941. At that time WPD perceived a need to warn its principals of a potential Japanese carrier-borne attack somewhere in the Pacific theater. For an account of the signals observed and background on how those signals were assessed, see, for example, the book by Wohlstetter (1962).

### 3.1. Illustration of the Crisis Definition Sub-Model

We define the dynamic states as the geographic position of the Imperial Japanese Navy (IJN) carrier strike force, a consolidated element of the combined fleet to which the Office of Naval Intelligence believed all Japanese aircraft carriers were assigned beginning on October 30, 1941. Specifically, we consider its position at a particular time, binned into discrete regions as described in detail in Section 3.2 and depicted in Figure 2, conditional upon our assessment of the strike force's next target and on the immediacy with which its orders direct it to proceed to that target. These two aspects of the strike force's intentions, target and immediacy, constitute the static joint variable D and are defined below. The analyst will reinitialize the sub-model if any significant development occurs that would change her prior assessment of the carrier strike force's target or the nature of its orders.

Figure 2: State space of pre-attack crisis representing possible locations of the IJN carrier strike force. Triangular areas as plotted on a Gnomonic projection have 648 nm sides. Outlined regions refer to trapping states for the possible targets (see further, in Table 3).

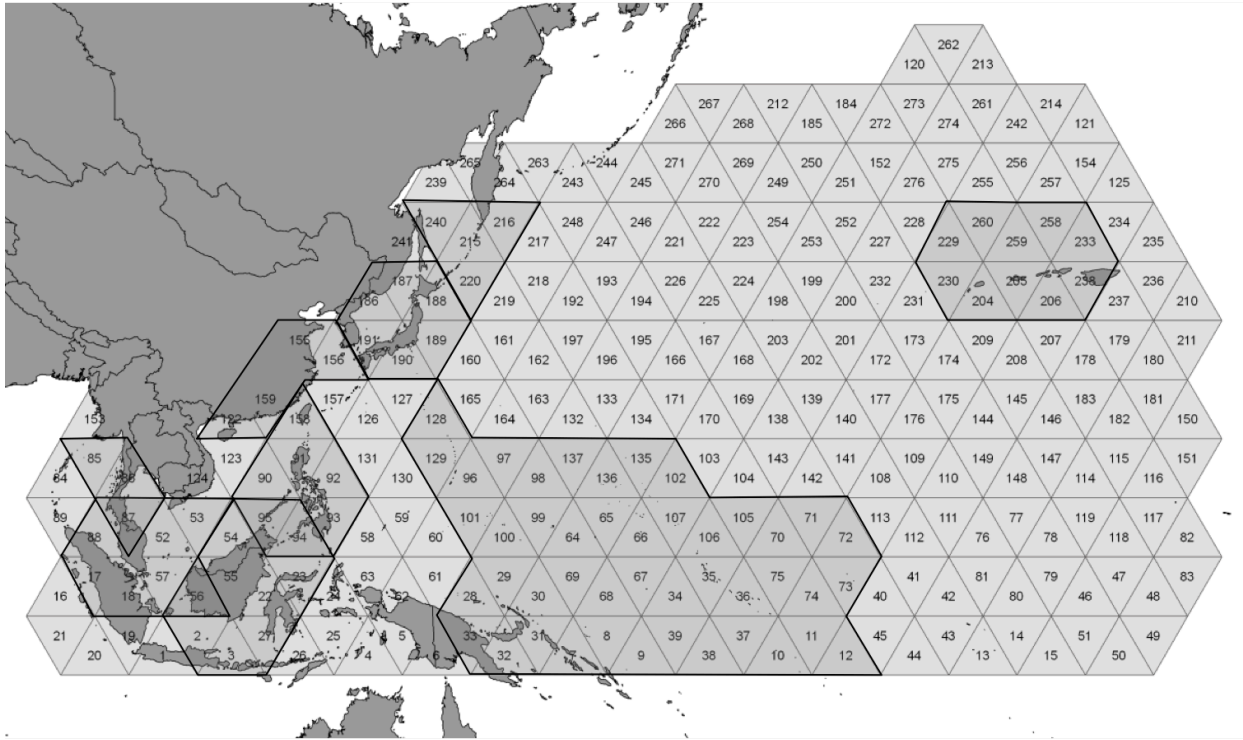

Figure 3: Influence diagram representing the Crisis Definition Sub-Model for warning of a Japanese attack (notations from Howard & Matheson 2005).

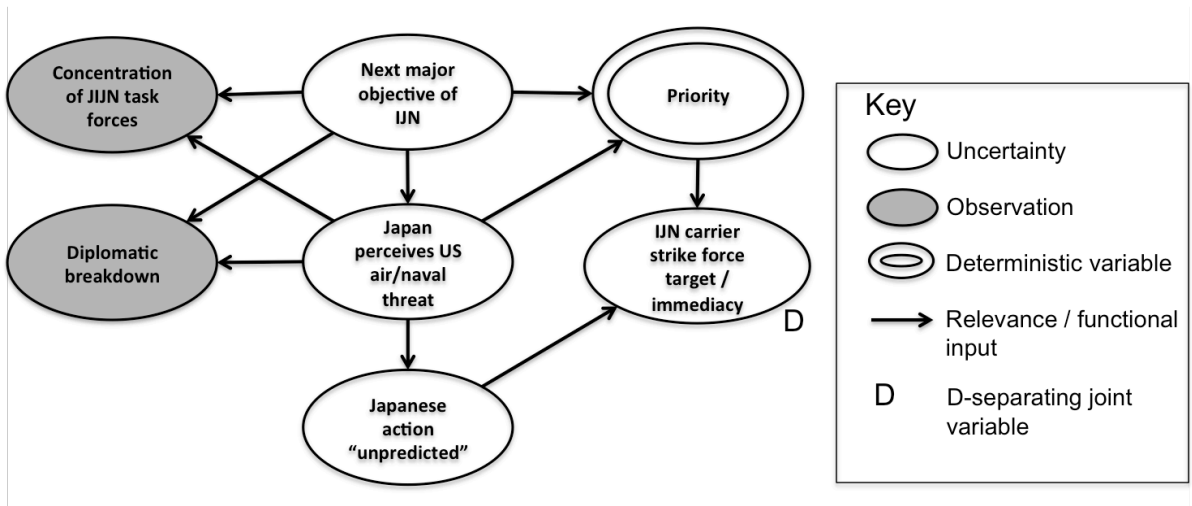

Figure 3 illustrates the Crisis Definition Model, depicting the IJN carrier strike force's uncertain intentions, given two observations and four intermediate variables. Based on an Army analysis from July 1941 addressed to WPD and echoed in subsequent WPD memos, we consider five possible "objectives" for the IJN carrier strike force, i.e. land masses that the Japanese high command might attempt to occupy using force projected from the carriers (US Army G-2 1946). Those objectives are MANILA BAY, SINGAPORE/KRA, BORNEO, THAILAND, and the KURILES (this report does not mention Hawaii). However, this and other intelligence reports and memos passing through WPD in November, 1941, cautioned that the Japanese might act "unpredictably", and in one report the term "unpredictably" is clearly used as a euphemism for attacking the US Pacific Fleet and nearby land-based aircraft, either in the Philippines or Hawaii. For that reason, we distinguish the IJN strike force's "objective" from its "target", i.e. land masses to which the IJN might be ordered to proceed and strike. Targets include the same five possible objectives plus OAHU. While the historical record validates US Navy assessments that the IJN *carrier strike force* was operating as a single unit, and as such, could not attack simultaneous targets, the IJN as a whole proved capable of exactly that, with some units operating without support from Japan's large-deck carriers. Therefore, this model illustration will be limited by the same failure of imagination as that of US naval intelligence in 1941, and focus exclusively on the whereabouts of the aircraft carriers based on their presumed ability to mount a single attack at a time.

In addition to "target," D (the underlying static fundamentals comprising the carrier strike force's intentions) includes two degrees of what we call "immediacy," or degree of urgency on the part of the Japanese fleet: IMMEDIATE and DELAYED. IMMEDIATE refers to orders to proceed to the target with haste. DELAYED refers to orders that would permit the carrier strike force to make port calls, steam at slow speed, or undertake other missions before attacking its specified target.

Other intermediate variables in Figure 3 include "Japan perceives US air/naval threat", "Japanese action unpredicted", and "Priority." Given an objective, WPD analysts were uncertainty whether the Japanese strike force commander would have perceived that the US Pacific Fleet and nearby land-based aircraft could threaten the conquest of his objective (YES or NO). We introduce a deterministic variable that we call "Priority" to simplify the conditional dependencies entering D. If "Japan perceives US air/naval threat" is NO, then "Priority" stores the same probability distribution as "Next major IJN objective", and if it is YES, "Priority" stores non-zero probability on MANILA BAY and OAHU, representing an attack on the US Pacific Fleet to defeat the threat. The target selected depends on which locations are contained in "Priority", along with "Japanese action unpredicted". All, except OAHU and MANILA BAY, are assigned a probability 0 if "Japanese action unpredicted" is YES. Otherwise OAHU is assigned a probability 0.

Lastly, based on their written assessments, two observations apparently influenced the beliefs of WPD analysts about Japanese strike force intentions at that time, a US-Japan diplomatic breakdown on November 24, and a report of the formation of two task forces concentrating in the South China Sea and the Mandates (US Chief of Naval Operations 1946a). A total of 103 assessments were performed in the Crisis Definition Sub-Model, although more than half were repetitions necessary to implement the logic of the PRIORITY deterministic variable.

### 3.2. Illustration of the Basic Warning Sub-Model

On a Gnomonic projection where great circles appear as straight lines, triangular rasters define possible positions of the Japanese carrier strike force and represent the state space **X**, as shown in Figure 2 in the preceding section. Their edge lengths are equal to the distance covered in two periods, and they cover the portion of the Pacific Ocean inside which the fleet might travel to strike any of its potential targets. A time unit is set as one-half day (12 hours). Assuming a speed of 27 knots, the edges of these triangular rasters are 27 x 12 x 2 = 648 nautical miles (nm) in length. We assume that the crisis will occur with certainty if and when the fleet arrives within striking range of its target, conservatively assessed to be 300 nm (US Cdr. Hawaiian Air Force 1946).

To specify the adjacency matrices $\mathbf{P_j}$ for each possible realization of D=j (target and immediacy), only transition probabilities between adjacent states are non-zero and must be assessed. Yet this still requires about 776 subjective assessments. A heuristic reduces the task, for example:

> The transition probability from state k to m given D=j is inversely proportional to an exponential of the shortest path from state m to the target (minimum number of rasters between m and the target).

This heuristic is an application of the concept of bounded rationality and is formalized in Equation 6. The carrier strike force is more likely, but not certain, to travel to states that are closer to its target. Implementing the following algorithm, which we refer to as the inverse-proportionality algorithm, allows the analyst to completely and automatically populate the transition matrices comprising **P** without performing any manual assessments except for the probability of holding (rasters including land masses that would block the passage of ships, such as rasters 85 and 86, are treated as non-adjacent).

1. Given state k and D=j, for all states m that are adjacent to k and for the set of states is set {k} that constitute trapping states (listed in Table 3), find the length of the shortest path (number of rasters) from m to any state in {k} and store as length(m)
2. $p_{jkm} \sim \frac{1}{\exp(length(m)+1)}$ (6)
3. Assess holding probabilities $p_{jk}$ and renormalize **P** accordingly.

Holding within a raster occurs if the strike force were to make a port call or slow its speed for other reasons. We assume a holding probability of 0.05 for all rasters given an IMMEDIATE target, or 0.4 for all rasters given a DELAYED target.[4] Using **$P_j$**, and **$\pi_t$** (as described in the next section 3.3), we calculate the probability distribution of the first hitting time $H_{\{k\},t}$ of the IJN carrier strike force to its target by applying Equation 1.

### 3.3. Illustration of Inference Sub-Model

On 26 November 1941 COM 16 sent a cable stating: "Our best indications are that all First and Second Fleet carriers still in Sase-Bo-Kure area" (US Cmdt. 16[th] Naval District 1946, p.16). ONI and G2 both interpreted this as indicating that the carriers were at Sase-Bo with certainty. Therefore, for **$p_0$**, a probability 1 is assigned to raster 191 and a probability 0 to all other rasters. In fact, November 26 was the last time where any radio signals were intercepted and attributed to the carrier strike force. Radio silence followed in all periods beginning on the morning of November 27 and ending with the attack of Pearl Harbor. As long as radio silence was not interpreted as white noise (equally likely for all locations of the strike force, its potential targets, and its level of urgency), it contained at least some new information.

We assess the likelihood of radio silence for each fleet position and objective and we use it to update **$\pi_t$** based on Equation 2, using a set of likelihoods that are consistent with a March 1941 study (US Cdr. Naval Base Defense Air Force & US Cdr. Hawaiian Air Force 1946), and with various other reports. Regarding an attack on Oahu, the approach under radio silence was deemed likely. The study authors, Major General Martin and Rear Admiral Bellinger, wrote, "It appears possible that Orange submarines and/or an Orange fast raiding force might arrive in Hawaiian waters with no prior warning from our intelligence service." In a subsequent report, MG Martin stressed that any force attacking the Hawaiian Islands would make a maximum effort to avoid detection in its approach (US Cdr. Hawaiian Air Force 1946). Yet, intelligence analysts in Hawaii and the Philippines in receipt of radio intercepts issued reports on several instances between November 27 and December 7 that read, "No indications of any movement any Fleet units," or, "Nothing to indicate Fleet out of home waters. " These analysts may have intended to express no assessment at all regarding fleet movement. If so, their vague language was misleading (US Cmdt. 14th Naval District 1994).

On the other hand, US analysts believed that a high degree of radio chatter was the norm in the Mandate islands. In the same report MG Martin and RADM Bellinger argued that the IJN would likely place a low priority on making a stealthy getaway following an attack, and would be willing to accept whatever casualties were necessary to ensure the success of a surprise attack. Radio silence would be lifted once combat operations were underway and operational security compromised. Applying this logic, we assess the conditional probabilities associated with radio silence and the validity of the observations in Table 3.

The US Navy possessed two independent sources for fleet radio intercepts in the West and Central Pacific, stations denoted COM14 and COM16, either of which may be functional or not. It is thus appropriate to define random vector R as comprised of two elements, each with two possible realizations, for a total of four possible realizations:

COM14 FUNCTIONAL, COM16 FUNCTIONAL
COM14 FUNTIONAL, COM16 NON-FUNCTIONAL
COM14 NON-FUNCTIONAL, COM16 FUNCTIONAL
COM14 NON-FUNCTIONAL, COM16 NON-FUNCTIONAL

Table 3: Coding of the probabilities of radio silence signals given the location of the IJN carrier strike force and at least one functioning antenna station, either COM14 or COM16.

| Location | Consistency of radio silence with | Probability of radio silence given | Rasters comprising locations {k}, see outlined regions in Figure 2 |
|---|---|---|---|

|  | US beliefs | location and functional station |  |
|---|---|---|---|
| near Philippines | partial | 0.5 | 90, 91, 92, 93, 94, 95, 158 |
| near Borneo | partial | 0.5 | 2, 3, 22, 23, 27, 54, 55, 56, 94, 95 |
| near Kra | partial | 0.5 | 17, 18, 52, 53, 56, 57, 86, 87, 88 |
| near Oahu | partial | 0.5 | 204, 205, 206, 207, 208, 209, 229, 230, 233, 238, 258, 259, 260 |
| near Kuriles | partial | 0.5 | 215, 216, 220, 240 |
| near Thailand | partial | 0.5 | 85, 86, 87 |
| near Mandate Islands and SE Pacific Ocean | inconsistent | 0.2 | 7, 8, 9, 10, 11, 12, 28, 29, 30, 31, 32, 33, 34, 35, 36, 37, 38, 39, 64, 65, 66, 67, 68, 69, 70, 71, 72, 73, 74, 75, 96, 97, 98, 99, 100, 101, 102, 105, 106, 107, 128, 129, 135, 136, 137 |
| near China | inconsistent | 0.2 | 122, 155, 156, 159 |
| home waters | consistent | 0.8 | 186, 187, 188, 189, 190, 191 |
| elsewhere in the Pacific & beyond range of targets | consistent | 0.8 | all others |

The Office of Naval Intelligence considered both sources to be reliable, but COM16 more so than COM14 (US ONI 1946c). Therefore, we assess the probabilities that COM14 and COM16 are reliable to be 0.7 and 0.9 respectively. In the case where both stations are not functional, we consider the probability that radio silence is observed from the strike force in any raster to be uniform, thus containing no information. In all, 276 x 4 = 1104 probability assessments are performed (most via automatic routine) in every time period.

### 3.4. Illustration of the Disutility Sub-Model

The Disutility Sub-Model assumes costs, risk attitude, and other preferences belong to a single principal decision maker. Until the 1986 Goldwater-Nicols Act, the US military operated under a split command paradigm whereby the Army, Navy, and later the Air Force each fought as three distinct forces. Therefore, in our illustration, we violate our own assumption regarding a single principal, since it would be wrong to consider either Admiral Stark or General Marshal as the sole principal. Incoherent decisions can arise in situations involving group decision making, and so too in war. The 1986 reform had the objective of fixing this situation, and by many accounts, it succeeded. Illustrations of the sub-models in the warning framework have, to this point, assumed probability assessments that are loosely based on the contemporaneous writings of the principals and their staffs. Unfortunately, even after reviewing the Foreign Relations of the United States and the papers and memos of senior officials in the White House, Army, Navy, and State Department, we find it is impossible to base an illustration of the Disutility Sub-Model on the beliefs and preferences expressed in such writings. Whereas one can justify (with several caveats) the discount rate that is used here, the costs assigned to the **q** vector and **V** matrix, and the lead times assigned to the **L** matrix, are not documented, and are justified only by the authors' sense of history.

The studies authored by MG Martin and RADM Bellinger on March 31, 1941, and by MG Martin on August 20, shed some light on perceptions of the costs of a missed alert and a false alert. In the former the authors wrote, "A successful, sudden raid, against our ships and Naval installations on Oahu might prevent effective offensive action by our forces in the Western Pacific for a long period." In the latter, MG Martin argued that Pearl Harbor's defenses were not impregnable, and that successful defense required finding the Japanese aircraft carriers before they launched an attack on Oahu (US Cdr. Naval Base Defense Air Force and US Cdr. Hawaiian Air Force 1946; US Cdr. Hawaiian Air Force 1946). It was also noted that moving the fleet out to sea would negatively impact training. We could not, however find any discussion of the relative costs of various Japanese operations against US interests versus placing US fleet on a high state of alert or moving its ships out to sea.

It is assumed that there is a unique alert type for every target. Table 4 shows alert costs and the costs of different outcomes in case of a Japanese strike in the absence of timely warning. Costs vectors **v** and **q** reflect some aggregate measure of military cost that incorporates lives lost, dollars spent, resources expended, and strategic loss. Because by that time war with Japan was considered inevitable, the total cost of war would not be "built into" the failure costs. Failure costs $v_{j\{k\}}$ are interpreted as the Joint Chiefs' willingness to pay some cost to avoid the expected losses arising from an attack launched by the IJN carrier strike force. $q_i$ is interpreted as the Joint Chiefs' willingness to pay some expected cost to avoid the specific measures they would undertake in response to a given warning. The cost of a successful strike on Oahu on November 27 is normalized to 1.

Table 4: Joint Chiefs' expected costs given target/alert type, normalized to attack on Oahu.

|  | $v_{j\{k\}}$ (failure cost of Japanese strike absent warning) | $q_i$ (expected cost of alert, including certain-equivalent of response) |
|---|---|---|
| Oahu | 1 | 0.00001 |
| Kuriles | 0.001 | 0.0000001 |
| Manila Bay | 0.1 | 0.00001 |
| Thailand | 0.0001 | 0.0000001 |
| Singapore/Kra | 0.01 | 0.000001 |
| Borneo/NEI | 0.01 | 0.000001 |

Table 5 shows the minimum lead times required to carry out interventions and avoid incurring failure costs $v_{j\{k\}}$, expressed in 12-hour periods (i.e. 2 periods per day). They reflect military plans to put ships to sea, bring in reinforcements, etc.

Table 5: Minimum lead times for alert type i to avoid incurring cost $v_{j\{k\}}$ in 12-hour periods.

|  | $l_{ij\{k\}}$, (minimum lead time for planned response) |
|---|---|
| Oahu | 4 |
| Kuriles | 1 |
| Manila Bay | 4 |
| Thailand | 1 |
| Singapore/Kra | 14 |
| Borneo /NEI | 14 |

We introduce a discount rate α to represent the Chiefs' perception that America would have been be better off the longer the outbreak of war was pushed off. It is impossible to disentangle *the general perception that delaying war would be good* because it would allow military forces to continue to train and prepare to fight (as discussed in the

Bellinger and Martin reports and associated memos), with *particular preparations* for war whose effect on the outcome of war is estimated in reports. We model the time varying costs using this constant discount rate, fitted using two points estimated by analyzing one specific thread of military preparations. We do this for three reasons: first, it was thought Japan might commit an act of war at any time; second, all reports and memos that commented on war preparation contained the opinion that later is better; and third, we believe that it is nearly as important that alerts that drive contingency operations not be premature as it is that they be timely. The Joint Board estimated that:

> "Strong diplomatic and economic pressure may be exerted from the military viewpoint at the earliest about the middle of December, 1941, when the Philippine Air Force will have become a positive threat to Japanese operations. It would be advantageous, if practicable, to delay severe diplomatic and economic pressure until February or March, 1942, when the Philippine Air Force will have reached its projected strength, and a safe air route, through Samoa, will be in operation (US Joint Board 1946).

In their November 27 memo to the President, General Marshal and Admiral Stark advised that delaying war until at least March 1942 would be advantageous. They mentioned reinforcements being sent east to the Philippines, including 21,000 troops whose ships would depart from the West Coast on 8 December. Los Angeles is 6339 nm from Manila, so at a relatively fast cruising speed of 13 knots, the reinforcements would have arrived on or about December 28. That date can serve as one anchor, with war any earlier perceived as costly, while war after about March 1, 1942, was perceived to have lesser cost, reflecting the effect of ongoing buildup and training of forces (Chief of Staff and Chief of Naval Operations 1946.) A daily discount rate between 1% and 2% compounded twice daily seems consistent with these statements. Normalizing the cost of war at the time of the decision to 1, with a 1% daily discount rate, the cost of war on December 28 has a present value of 0.73, and on March 1, 1942, of 0.39. With a 2% discount rate, those present value costs are 0.54 and 0.15 respectively. A daily discount rate of 1.5% is used further in this paper, thus splitting the difference between these present values.

For simplicity, the principals' expected disutility curve is assumed to be linear (risk neutral). Consistent with the illustration of the Inference Sub-Model in Section 3.3, the analyst observes radio silence in each period, which is incorporated into the analyst's belief regarding the current state of the IJN carrier task force. In every period beginning November 27, 1941, the analyst will use her latest beliefs regarding D (the adversary's capabilities and intentions) and the current state of the strike force to scan ahead using the three-step procedure described Section 2.

### 3.5. Illustration's Analytic Results
### 3.5.1. Illustrative Results of the Crisis-Definition Sub-Model

Table 6 shows the prior probabilities associated with the IJN's next major objective and carrier strike force target at the end of November, as well as the posterior probabilities after observing the diplomatic breakdown and task force formation. The priors reflect the authors' assessment of the subjective beliefs held by the principals based on the documents previously cited. The probability that a location is the next target is distributed evenly between IMMEDIATE and DELAYED degrees of urgency. While the Joint Chiefs' posterior beliefs that the IJN strike force would target US forces should have exceeded 94 percent by the inferential logic described above and using our probability assessments, it appears that the analysts working for them did not perform this updating since successive analyses still identified Thailand, Malaya, or the Netherlands East Indies as most likely Japanese targets.

Table 6 Probabilities (in percentages) of IJN objective versus carrier strike force target.

|  | IJN's next major objective | | IJN carrier strike force target | |
| --- | --- | --- | --- | --- |
| Objective / target | Prior | Posterior | Prior | Posterior |
| OAHU | N/A | N/A | 6.2 | 9.51 |
| KURILES | 15 | 0.069 | 14.5 | 0.1 |
| MANILA BAY | 15 | 25.7 | 55.3 | 85.5 |
| THAILAND | 30 | 12 | 17.6 | 1.61 |
| SINGAPORE/KRA | 15 | 23.3 | 2.49 | 1.24 |
| BORNEO | 25 | 38.9 | 3.85 | 1.99 |

### 3.5.2. Illustrative Results of the Basic Warning Sub-Model

Figures 4 and 5 represent the first passage time distributions $H_{k,t=Nov.\ 27}(\tau)$, obtained by applying Equation 1 to the vector of transition matrices **P** specified in Section 3.2, for the time at which the IJN carrier strike force arrives within striking range of its target under IMMEDIATE and DELAYED conditions. The series of zero probabilities at the beginning of each graph reflects the fact that too little time has elapsed to permit the Japanese fleet arrive at its striking range of each possible target. Using the results of Section 3.5.1 to marginalize the strike force's target and immediacy, Figure 6, shows the joint cumulative distribution over the target and the first passage time into the various crisis states.

Figure 4: Cumulative conditional probability $H(\tau)$ that the IJN carrier strike force reaches its target's location given each target and IMMEDIATE mode. Authors' assessments based on information available on Nov. 27, 1941.

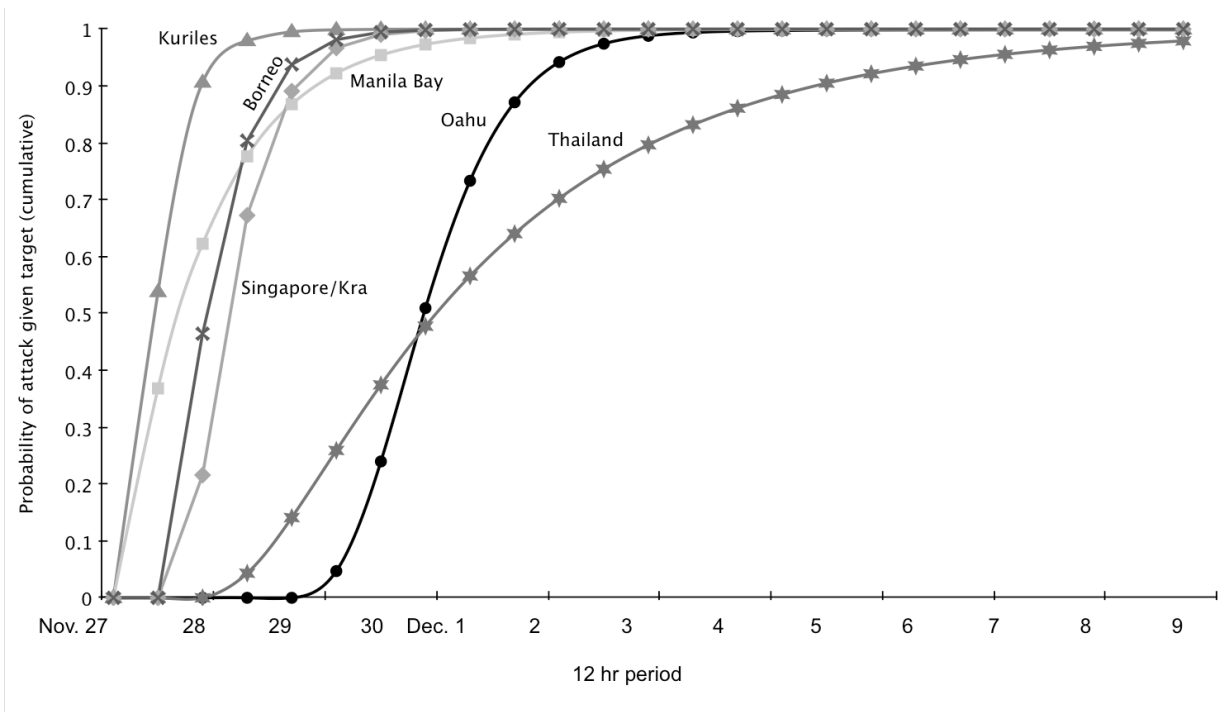

Figure 5: Cumulative conditional probability $H(\tau)$ that the IJN carrier strike force reaches its target's location given each target and DELAYED mode. Authors' assessments based on information available on Nov. 27, 1941.

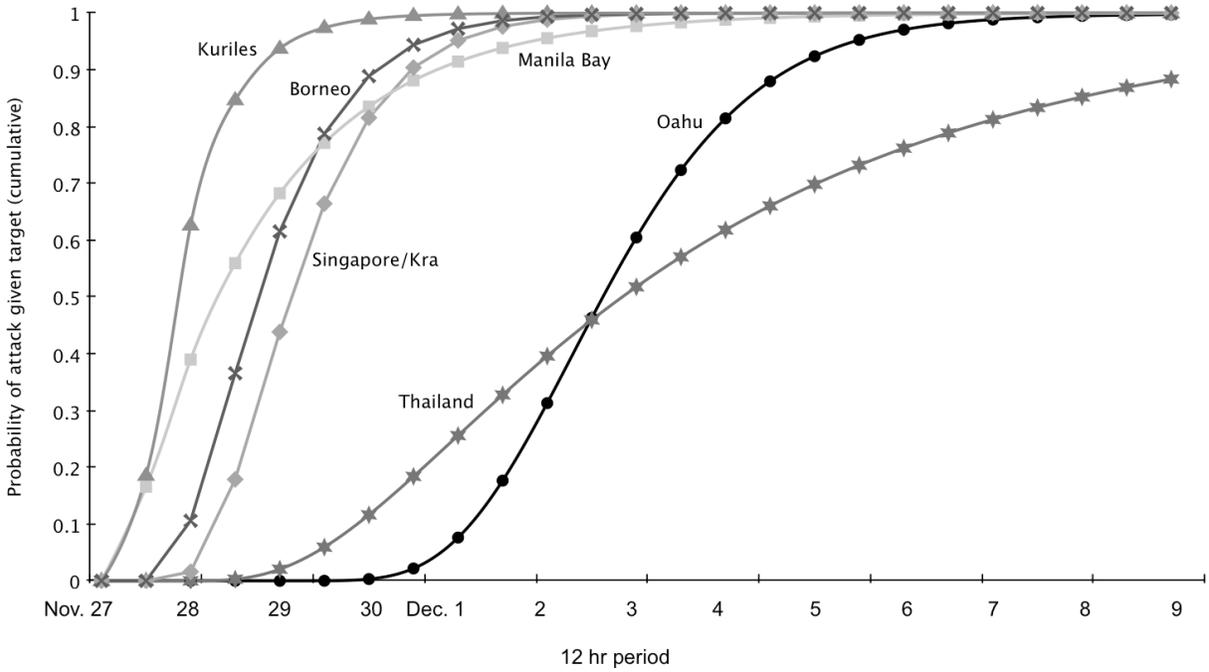

Figure 6: Cumulative joint probability distribution over the IJN carrier strike force's target and time it reaches its target's location, in half-day time periods beginning Nov. 27, 1941.

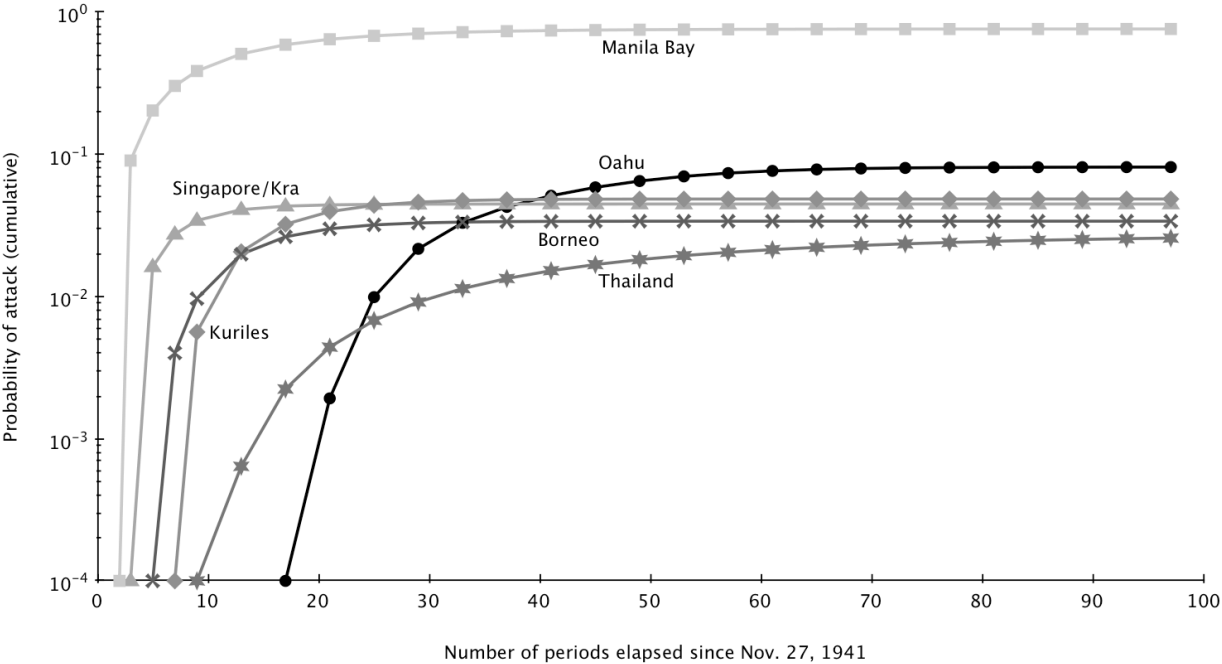

These distributions, if they had been available, could have conveyed to the principals on November 27, 1941, the likelihood that Japan strike each of the six targets on or before any given 12-hour period in the future. If RADM Turner, then director of the Navy's War Plans Division, had told his superior, Admiral Stark, "It will take 30 days for the Army to send reinforcements to the Philippines, and without those reinforcements the Asiatic Fleet is gravely vulnerable," Admiral Stark could have looked at the probability of an attack on Manila at t=60 time units, and decided whether he was willing to accept the probability of 0.85 that the Japanese would have struck by that date. Courses of action might have included ordering the Asiatic Fleet to sea. Similarly, the distribution corresponding to an attack on Oahu could have helped a principal in deciding whether to intervene, for instance by ordering the US Pacific Fleet to sea, based on an estimate of the time required for a delivery of the full complement of B-17 bombers that MG Martin believed he would need to adequately secure Oahu from a Japanese surprise attack.

### 3.5.3. Illustrative Results of the Inference Sub-Model

The analysis of the Inference Sub-Model was made using Murphy's Bayesian Network Toolbox, written for Matlab (Murphy 2002). Distributions similar to those shown in Figures 4, 5, and 6 were computed for the beginning of each half-day time period between November 27 and December 6 (inclusively), where in each time period, $\pi_t$ is updated by a RADIO SILENCE signal. Results are shown in Figures 7 through 10. Figures 7 and 8 illustrate how an analyst's belief regarding the D-separating joint variable (the IJN carrier task force's target and sense of urgency) is updated following each radio silence observation. For clarity, the data are displayed as two figures instead of one.

Figure 7: Probability that the IJN strike force is moving towards a target in an IMMEDIATE mode given radio silence is observed in each time period.

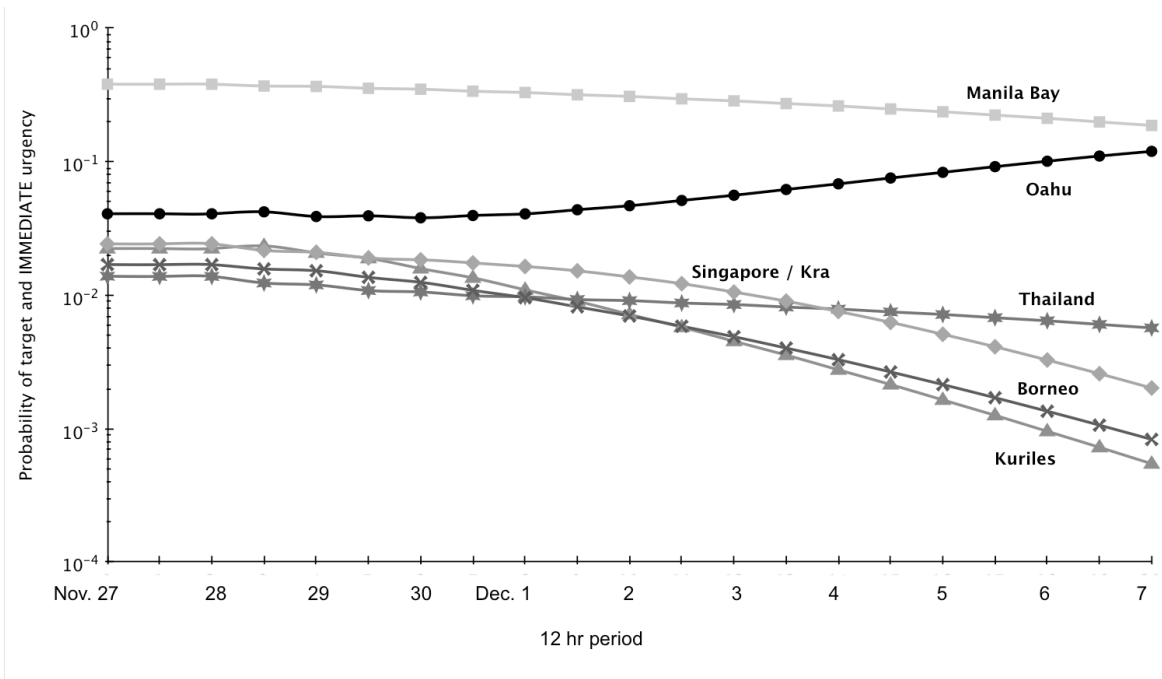

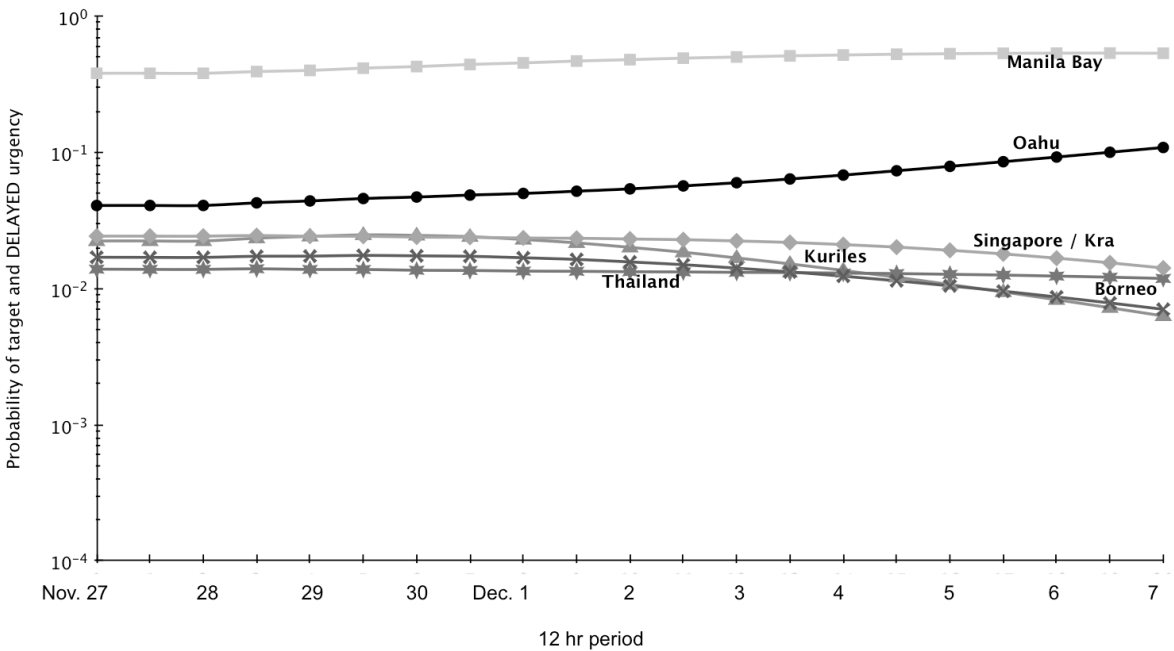

Figure 8: Probability that the IJN strike force is moving towards a target in a DELAYED mode given radio silence is observed in each time period.

According to the Office of Naval Intelligence reports published on December 1, intelligence analysts had concluded that the IJN carrier task force was still in home waters (US ONI 1946a, 1946b). This conclusion is consistent with the increasing likelihood that

Manila Bay was a DELAYED target and with increasing likelihood that Oahu was an IMMEDIATE target.

Figures 9 and 10 apply Equation 3 to illustrate how the analyst would have made projections ahead in time after updating her beliefs based on observed radio silence beginning on November 27. They show 2-day, 4-day, and 7-day projections for an attack on Oahu and Manila Bay respectively. The former shows low probabilities of a future attack on Oahu, even based on a 7-day projection, although the probabilities increase with each passing period (still 2.5% by December 6). The latter shows significant probabilities of an attack on Manila Bay, even projecting ahead only two days into the future. Those probabilities peak in the first period (evening November 27) and steadily drop with each passing day because the longer radio silence persists, the more likely it is estimated that the strike force has gone elsewhere.

Figure 9: Probability of an attack on Oahu 2 days, 4 days, and 7 days after each period beginning November 27, 1941, given radio silence observed prior to each period.

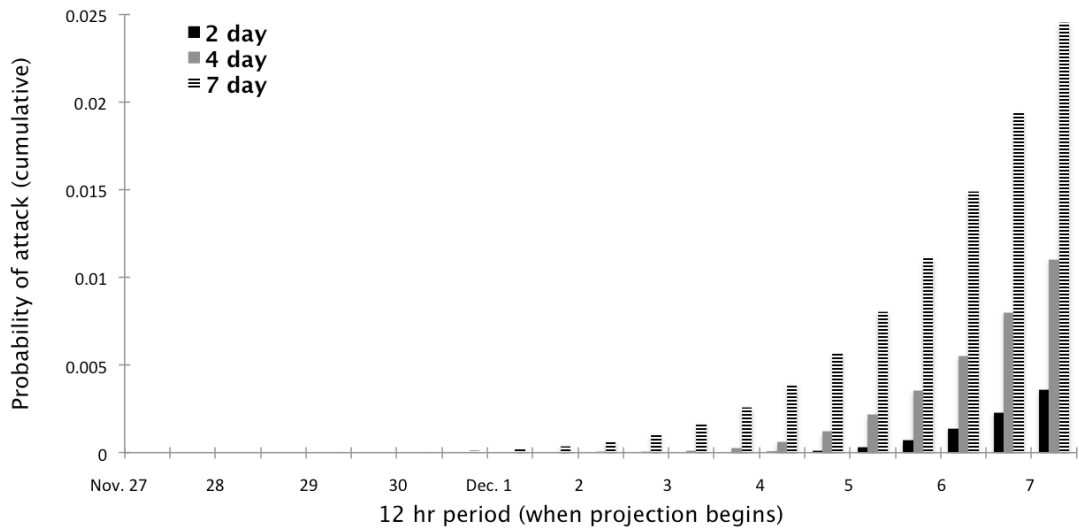

Figure 10: Probability of an attack on Manila Bay 2 days, 4 days, and 7 days after each period beginning November 27, 1941, given radio silence observed prior to each period.

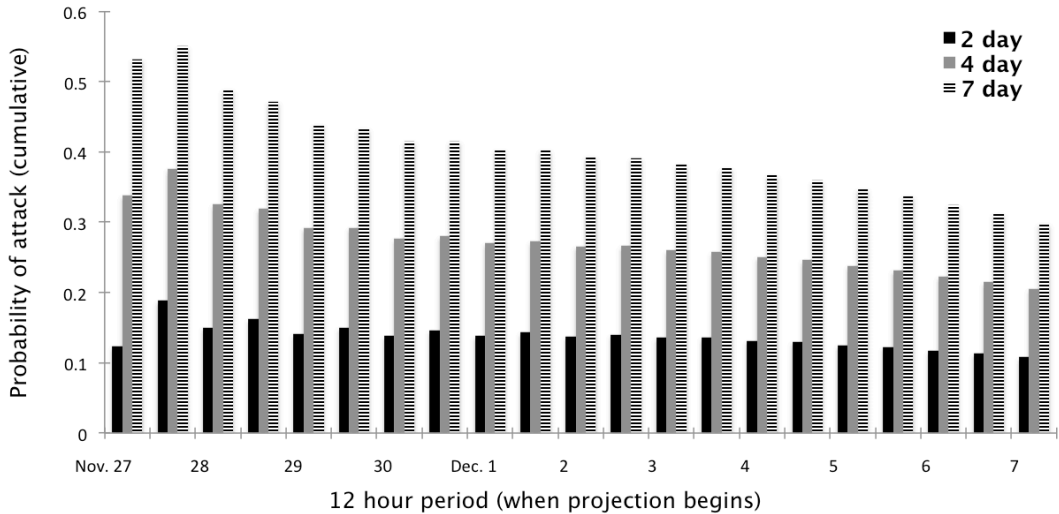

### 3.5.4. Illustrative Results of the Disutility Sub-Model

Applying Equation 4 to find expected disutilities yields results shown in Figures 11 and 12 (note that smaller expected disutilities are displayed above larger expected disutilities on the vertical axis). Using our illustrative probabilities and costs, the model generates an alert for an attack on Oahu on December 2, thirteen periods after the model's initialization.

Figure 11: Scan-ahead on November 27, 1941, for the expected disutility of an alert.

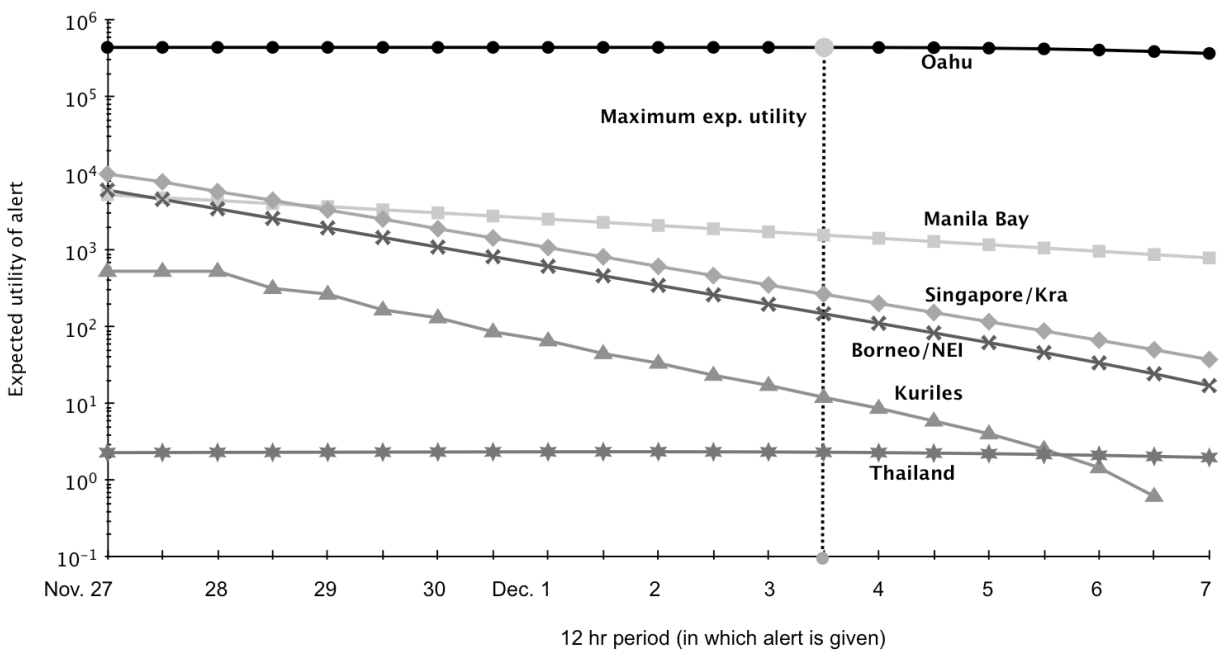

From Sase-bo, Japan (raster 191), the minimum transit time implied by **P** to Oahu is 9 periods, or 4.5 days, as represented in Figure 4. Recall the minimum lead time that we assume for an attack on Oahu is 4 periods, or 2 days. On December 2, the probability inferred from radio silence of an attack on Hawaii does not even register on Figure 9! However, the extreme expected cost that we have adopted implies that an alert for an attack on Oahu on that date is optimal. Beginning on November 27, the very slight increase in expected disutility with successive periods (until the 13th period), as seen in Figure 11, occurs because the cost of warning is discounted for every period where the analyst waits to issue an alert. Given the time of the crisis, the Disutility Sub-Model reflects no benefit to a warning that is earlier than the minimum lead-time. As one might expect, with each passing period, after updating by inference given the radio silence, the minimum expected disutility slides one period closer to the present and remains associated with an Oahu alert. We discuss the sensitivity of this alert to the various probabilities and costs in the next section.

Figure 12: Scan-ahead showing expected disutility of an alert on December 2, 1941.

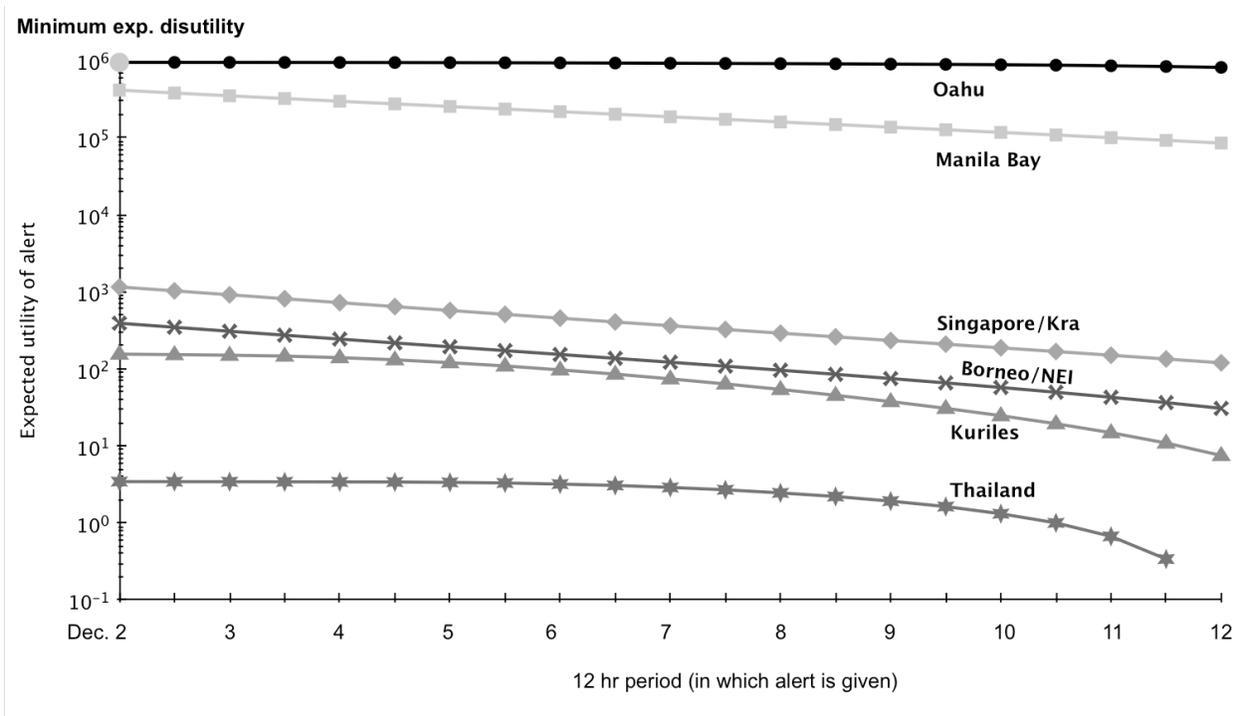

Again, the analysis presented in this section is based on a past known crisis and is entirely illustrative. Only the slightest evidence exists to suggest that US principals or their staffs would have held the values and preferences we used as inputs. The objective is to present a framework for information gathering and decision reasoning, understanding that we cannot know now what values the principals held at that time. Clearly, the decision to warn of an attack on Oahu on December 2, like the results of a typical decision model, is strongly sensitive to the expected costs associated with each potential outcome and necessary lead time (i.e. v, q, and l). Within certain intervals for two of these three parameters, the model will be highly sensitive to the third input, and in other intervals, it is not. Around the input values used in this illustration, the model was more sensitive to failure costs, but less sensitive to fixed alert costs and the necessary lead times. That said, if one accepts that the fixed costs associated here with an alert, and the failure costs associated with attacks on the various targets are roughly in the right orders of magnitude relative to one another, an attack on Oahu, which has considerably lower probability than an attack on Manila Bay in any time period, is shown to be of much greater concern.

### 3.6. Discussion of Illustrative Results

Figures 4 and 5 provide a window into the modeling framework's sensitivity to transition probabilities. Whether the IJN carrier strike force is in an IMMEDIATE versus DELAYED mode of urgency affects the probability that it remains in the same raster for two consecutive time periods (0.05 for IMMEDIATE, 0.4 for DELAYED). The latter implies that attacks on all potential targets are slower to develop. This does not substantially affect the rate at which the likelihood of an attack increases on nearby targets (e.g. Manila Bay), but it does impact the rate at which the likelihood of an attack increases on a distant target (e.g. Oahu). Given that Oahu is an IMMEDIATE target, on November 27 the likelihood of an attack on or before December 11 (period 30) exceeds 0.57, while it is just 0.03 if Oahu is considered a DELAYED target. In other words, the Basic Warning Sub-Model is quite sensitive to this "holding probability", and to the transition probabilities in general. This sensitivity, in fact, led to our decision to employ an inverse-proportionality heuristic that depended on the exponential of the remaining time of travel, rather than simply the remaining time of travel. The alert of an attack on Oahu discussed in Section 3.5.4 would have occurred after December 2 given a higher "holding probability" (especially in the IMMEDIATE operating mode). However, the operating modes are intended to reflect the nature of orders given to the commander of the strike force. It seems difficult to imagine that WPD would have assigned its adversary any significant likelihood of staying put in one place for 12 hours while under orders to proceed directly to attack a target.

The framework is also sensitive to the initial distribution over the D-separating joint variable (the INJ carrier strike force's target and sense of urgency). The posterior distribution of D that is shown in Figures 7 and 8 places far more likelihood on an attack against Manila Bay than it does on Thailand (in period 1, 0.885 on Manila Bay and 0.016 on Thailand). While these initial values, which were derived from the Crisis Definition Sub-Model described in Section 3.1, drove all subsequent probability distributions of an attack, a uniform distribution over D would have produced a somewhat different probability distribution over the time and place of a future attack. The distribution on D constructed in the sub-model yielded an attack on Manila Bay on or before December 11 (period 30) with a 0.71 probability and on Thailand with a 0.01 probability. A uniform distribution would have yielded 0.16 and 0.06 respective probabilities of attack by December 11. Although the extreme disparities in the probabilities of attacks shown in Figure 6 follow from this particular distribution over D, even in the case of a uniform initial distribution, there would still have been a 2 ⅔ fold disparity.

By contrast, while a signal that is assessed to express near-certainty will drive the results produced by the Inference-Sub-Model, these results are rarely sensitive to uncertainties in signal assessments given the state of the crisis. To illustrate this point, suppose that COM16 had intercepted a radio signal from the carrier strike force on December 1, indicating that it was more likely to have originated from the northern Pacific than the South China Sea or elsewhere. The projections corresponding to probabilities of attacks on Oahu and the Kuriles would have increased and the projections for attacks elsewhere would have decreased to some degree. However, subsequent observations of radio silence would have restored the probability projections that preceded a single radio intercept because they would have informed an analyst that the strike force was moving toward its target. This means that prior beliefs still contribute substantially to the updated probability. Nonetheless, a steady sequence of such signals, or a single signal that is assessed to be highly diagnostic and thus overwhelming the priors, would have a great impact on the results. Analysts should thus avoid attributing certainty to signals, and expressions of near-certainty should be subjected to scrutiny by "red-team" analysis. By the same token, had we assumed that radio silence had a probability less than 0.8 given "all other locations" than those listed in Table 3, inference that in each successive period the fleet was increasingly likely to attack Oahu would have proceeded more slowly, again delaying an alert.

In reality, the Office of Naval Intelligence interpreted the radio silence as an indication that the carriers were in their homeports (US ONI 1946b) and did not sufficiently considered other possibilities. This is a result of what Tversky and Kahneman (1974) called "availability bias" and which has become popularly referred to as "confirmation bias". Assuming that our analytic framework had been in use, this analysis and in particular the results from Figures 9 and 10 suggest that an analyst would not have put a high probability on a Japanese surprise attack on Pearl Harbor by December 7, 1941 precisely because the prior for an attack on Oahu was so low. It is this prior that accounts for the substantial discrepancy between reality and the posterior probabilities of an attack on Oahu and on Manila Bay after observing days of radio silence. Based such observation, the posteriors slowly approach each other and might eventually cross, but with priors that differ so substantially from the reality of Japanese intent, this would take substantially longer than 20 periods (10 days). These prior probabilities represent the authors' best efforts to reflect the state of information of US principals and analysts expressed in writing in the historical records. That information was clearly incomplete.

One therefore sees both the value and the limits of a probabilistic warning model. It allows an analyst to make inferences based on noisy signals and to project the results ahead to generate warnings, bringing consistency in the incorporation of new information to generate posterior probabilities. However, it will never inform an analyst that those probabilities (both the priors and the signal likelihoods) are incorrect. If the analyst has a prior belief that a Japanese attack on Manila Bay is far more likely than an attack on Oahu, it will be carried forward in the analysis. What this model does is to provide a method of systematically processing information, but it cannot assess the accuracy of the input.

The key to understanding the results is that warnings must involve the prospects associated with different scenarios in a crisis and not be restricted to the most likely hypothesis. Decision makers' actions, and those of their analysts, will reflect the costs they associate with each possible outcome and the other priorities that they have to deal with. General Marshal and Admiral Stark are likely to have realized in the Fall of 1941, that a Japanese surprise air attack on Pearl Harbor would be absolutely devastating to the US Pacific Fleet, and by extension to the United States as a nation. Distributions of the sort shown in Figure 9 could have highlighted, for instance, that on December 5 even a ½ percent chance that the IJN carrier task force would attack the US Pacific Fleet at anchor within 4 days posed a significant risk, especially in light of the increasing trend. Depending on whether they thought that the relative cost of ordering the fleet to sea was low enough, maybe it would have been worthwhile despite the small probability of an attack on Oahu.

As widely recognized, risk cannot be measured by probability alone, and the costs associated with crisis outcomes, true alerts, false alerts, and missed alerts, are integral elements of the value of a warning. Someone must be responsible for integrating these factors. Traditionally the responsible individual is the policy maker, but given she expects warnings from her analyst, it is clear that the analyst must help in this integration in order to fulfill the warning task. This will work provided that she understands and can adopt the preferences of the decision maker, and also properly communicates uncertainties in the priors and in the evaluation of the signals.

Figures 11 and 12 show the effects of both errors of Type I (missed alert) and Type II (false alert) on an analyst's decision. The costs of Type I errors drive the low expected disutility of warning of an attack on OAHU despite its low probability. On the other hand, because the principal is assumed to discount all future costs to their present value at a constant discount rate, the fixed cost associated with an earlier warning (true or false) may be greater than that of a later one. This fixed cost associated with both false and true alerts provides an incentive to the analyst to postpone a warning until he or she has some confidence that the alert has merit; but it should not be to the point that it misses the lead time requirements and inflicts the cost of missing an alert. The tipping point was shown in this illustration to have occurred on December 2.

Finally, considering only the final model result, any given warning recommendation may seem opaque. Viewed separately, however, the results of the Crisis Definition Sub-Model, Inference Sub-Model, Basic Warning Sub-Model, and Disutility Sub-Model are more intuitive, and the sub-models are substantially more transparent individually and for a single time unit than collectively. For example, in the Inference Sub-Model, we found that changing the likelihood of radio silence given various possibilities at a given time yielded intuitive changes in the probability distributions of the present state of the crisis and of the IJN carrier task force's target. As in any complex argument, effective communication of a model's results to a stake-holder requires breaking that argument into its components, in this case, each of the sub-models and their results.

### 3.7. Discussion of Implementation

Our model considers a single analyst who works directly for a single principal and is able to communicate an alert without impedance of any kind. While this arrangement accurately represents how certain principals communicate with their analysts – typically principals in command of tactical echelons – it may not represent realities in many organizations. As in all efficient "organizational warning systems" (Lakats and Paté-Cornell 2004), communications from individual analysts to principals must be screened through multiple layers of review. Absent this arrangement, national-level principals would be overwhelmed by communications from throngs of analysts.

According to the Defense Strategies Institute (2014), "With ever increasing technology and sophistication of sensors, the amount of information being collected has overwhelmed analysis systems and current processes." We suggest that one of the keys to making this model useful in the real world will be encoding early assessments of entire classes of signals that are common to different types of crises, or at least grouping them by sets of features characteristic of certain classes of signals. These classes of signals that are observed repeatedly should be characterized according to their key technical and circumstantial features. Each signal class, given its main features, can then be assessed according to the extent to which it conveys the presence or movement of certain classes of entities, from which one can derive a signal's likelihood. This categorization will avoid the need for an analyst to perform a set of assessments for each individual observation from the myriad of sensors that the US intelligence community deploys. Instead, the system allows integration of data in bulk, which we believe is what Ward Edwards et al. (1968) envisioned nearly 50 years ago. The feature identification and assessment, however, still involves important analytic judgment. No amount of "pre-assessment" will replace a good analyst in performing early warning analysis.  Additionally, "pre-assessment" could increase the risk already present in the model that the crisis dynamics may not be wholly identified and defined *a priori*. Alerts are generated only with respect to crises that have already been imagined and are reflected in the Basic Warning Sub-Model's state space, again making the model more appropriate for tactical than strategic warning.

Several authors have written that the chief drawback of Dynamic Bayesian Networks and Partially Observable Markov Decision Processes is the "curse of dimensionality." In our Pearl Harbor example, we performed a total of 12 x 1104 = 13,248 assessments in the Inference Sub-Model before arriving at a decision to warn on December 2, along with 103 assessments in the Crisis Definition Sub-Model and 776 assessments (derived from a heuristic) in the Basic Warning Sub-Model. Twelve values were assessed in the Disutility Sub-Model.  These 14,000-odd assessments do indeed reflect a curse of dimensionality. But they do not have to be performed for each time unit, and many of these data are likely to remain unchanged after an initial assessment. Therefore, the number is not so large as to make this analysis infeasible in light of the heuristics that were used to derive some probabilities and the repetitive nature of generating others. Still, this data requirement does convey the importance of assessing conditional probabilities in bulk, both to reduce the workload to something manageable and to reduce the data demand to a level that would not overwhelm a principal decision maker trying to obtain insight from the model. The alternative is to analyze the crisis across fewer dimensions, with the increasing possibility of missing an important factor. The ultimate alternative is a mere seat-of-the-pants assessment and decision, reflecting experience but perhaps missing some logic and new circumstance.

## 4. CONCLUSIONS

In this paper, we argued in favor of using dynamic, probabilistic (Bayesian) reasoning to perform early warning analysis. We presented a formal approach for doing so, using iterative probabilistic inference, projection in time, and decision analysis. Maintaining this kind of model may be time-consuming. Therefore, we envision an early warning system where individual analysts can use computer software to assist them in structuring their views of the nature of evolving threats, in addition to entering probability assessments derived from intelligence information in near-real time.

Some principals may view with some degree of distrust the kind of modeling results presented to them as if from a black box. This, however, is no different from how they might have felt about assessments derived from traditional *indications & warning* that omit discussion of trends (Belden 1977). The system will only be as credible as the analyst's mental model of evolving threats and her willingness to consider carefully the implications of these mental models through the system. Given that willingness, our model offers an advantage over traditional *indications & warning* methods in that it consistently applies an analyst's logic with regard to crisis dynamics, regardless of the amount of intelligence information.

The illustration presented here is based on a historic national security (and geographic) problem. Crisis dynamics and warnings are also important in many other domains such as medicine, or in business and finance such as when facing a propagation of failures among globally connected banks. Processing information about signals to estimate the progress of a crisis and to evaluate the need to warn a principal of an impending disaster allows analysts, nurses, or bankers to structure and strengthen their mental models for a wide variety of crises and possibilities. Analysts, and anybody close enough to the system to observe its evolution, will then be free to spend their time assessing signals (their meanings and their dependences) and updating priors, with confidence in their logic and consistency.

## NOTES

1. Formally, the Markov Property may be written as:

$$Prob\{X_{t+1} = x_{t+1} | X_t = x_t\} = Prob\{X_{t+1} = x_{t+1} | X_t = x_t, X_{t-1} = x_{t-1}, \ldots, X_0 = x_0\}$$

The state space **X** may be expanded to incorporate any finite amount of historical memory.

2. For an analysis of a real-world crisis warning problem involving a transportation network used to define the state space, see Blum (2012).

3. Assessing source validity is beyond the scope of this research. The basic method of intelligence source reliability assessment can be found in Chapter 12 of Field Manual 2-22.3 (U.S Army 2006) and in Intelligence Analysis: A Target Centric Approach (Clark 2009).

4. Each transition matrix $P_j$ requires 85 pages of description, and there are 12 such transition matrices corresponding to the 12 realizations of the D-separating joint variable. They are not included here but will be made available electronically upon request.

## ACKNOWLEDGEMENTS

This research was supported in part by the United States Department of Homeland Security through the National Center for Risk and Economic Analysis of Terrorism Events (CREATE) under sub-award number 46893. Any opinions, findings, and conclusions or recommendations expressed herein are those of the authors and do not necessarily reflect the views of the Department of Homeland Security or the United States Government.

## REFERENCES

Abbas AE, Matheson JE (2005) Utility transversality: a value-based approach. Journal of Multi-Criteria Decision Analysis, 13(5-6): 229–238.

Baum LE, Eagon JA (1967) An inequality with applications to statistical estimation for probabilistic functions of Markov processes and to a model for ecology. Bulletin of the American Mathematical Society, 73(3):360–363.

Baum LE, Petrie T (1966) Statistical inference for probabilistic functions of finite state Markov chains. Annals of Math. Stat., 37(6):1554–1563.

Belden TG (1977) Indications, warning, and crisis operations. International Studies Quarterly, 21(1):181–198.

Bellman RE (1953) An introduction to the theory of dynamic programming. RAND Corp., Santa Monica, Calif.

Blum, DM (2012) Probabilistic models for warning of national security crises. Ph.D. dissertation, Dept. Management Sci. and Eng., Stanford University, Stanford, Calif.

Clark R (2009) Intelligence analysis: A target-centric approach, 3rd ed. CQ Press, Thousand Oaks, Calif.

Defense Strategies Institute (2014) Big data for intelligence symposium. http://bigdatasymposium.dsigroup.org/

Edwards W, Phillips, LD, Hays WL, Goodman, BC (1968) Probabilistic information processing systems: Design and evaluation. IEEE Trans. Sys. Sci. and Cybernetics, 4(3): 248-265.

Geiger D, Verma T, Pearl, J (1990) d-separation: From theorems to algorithms. Proc. 5th Conf. Uncertainty in Artificial Intelligence, Windsor, Ontario. 139–148.

Grabo C (1994) Strategic warning: The problem of timing. Central Intelligence Agency Historical Review Program. https://www.cia.gov/library/center-for-the-study-of-intelligence/kent-csi/vol16no2/html/v16i2a07p_0001.htm

Heuer Jr RJ (1981) Applications of Bayesian inference in political intelligence. Hopple GW, Kuhlman JA, eds. Expert-generated data: applications in international affairs. Westview Press, Boulder, Colo.

Howard RA (1960) Dynamic programming and Markov processes. Technology Press of MIT, Cambridge, Mass.

Howard RA (1989) Knowledge maps. Management Sci., 35(8): 903-922.

Howard RA, Matheson JE (2005) Influence diagrams. Decision Anal., 2(3): 127–143.

Lakats LM, Paté-Cornell ME (2004) Organizational warnings and system safety: A probabilistic analysis. IEEE Trans. Eng. Management 51(2): 183-196.

Murphy KP (2002) Dynamic Bayesian networks: Representation, inference and learning. Ph.D. dissertation, Dept. Computer Sci., University of California, Berkeley, Calif.

Paté-Cornell ME (1986) Warning systems in risk management. Risk Anal., 6(2): 223–234.

Paté-Cornell ME, Fischbeck PS (1995) Probabilistic interpretation of command and control signals: Bayesian updating of the probability of nuclear attack. Reliability Eng. and System Safety, 47(1): 27-36.

Peterson WW, Birdsall TG, Fox W (1954) The theory of signal detectability. Trans. IRE Professional Group on Info. Theory, 4(4):171-212.

Shachter RD (1998) Bayes-ball: The rational pastime (for determining irrelevance and requisite information in belief networks and influence diagrams). Proc. 14th Conf. Uncertainty in Artificial Intelligence, San Francisco, Calif.

Schelling TC (1960) The strategy of conflict. Harvard University Press, Cambridge, Mass.

United States. Asst. Chief of Staff, G-2, US Army (1946) Ex. 33, Memorandum for Chief of Staff, WPD, July 11, 1941. Hearings before the Joint Committee on the Investigation of the Pearl Harbor Attack (hereafter Hearings). GPO, Washington, DC.

United States. Army (2006) Field manual 2-22.3 Human intelligence collector operations. GPO, Washington, DC.

United States. Chief of Naval Operations (1946) Ex. 144, Cable 242005, November 24, 1941. Pearl Harbor attack: Hearings. GPO, Washington, DC.

United States. Chief of Staff, US Army, and Chief of Naval Operations, US Navy (1946) Ex. 17, Memorandum for the President, November 27, 1941. Hearings. GPO, Washington, DC.

United States. Commandant, 14th Naval District (1994) Traffic intelligence summaries, July-December, 1941, SRMN-012. Parker F, ed. Pearl Harbor revisited: United States Navy communication intelligence, 1924-1941. Center for Cryptologic History, Ft. Meade, MD.

United States. Commandant, 16th Naval District (1946) Ex. 8, Cable 261331, November 26, 1941. Hearings. GPO, Washington, DC.

United States. Commander, Hawaiian Air Force (1946) Ex. 13, Study of air situation in Hawaii, August 20, 1941. Hearings. GPO, Washington, DC.

United States. Commander, Naval Base Defense Air Force & United States. Commander Hawaiian Air Force (1946) Ex. 144, Joint estimate covering Joint Army and Navy air action in the event of sudden hostile action against Oahu or fleet units in the Hawaiian area, March 28, 1941. Hearings. GPO, Washington, DC.

United States. Commander US Pacific Fleet, 1946. Ex. 44, Pacific Fleet Confidential Letter 2CL-41 (Revised), October 14, 1941. In Hearings. GPO, Washington.

United States. Joint Board (1946) Ex. 16, Minutes of meeting, November 3, 1941. Hearings. GPO, Washington, DC.

United States. Office of Naval Intelligence (1946a) Ex. 149, Fortnightly summary of current national intelligence, Serial no. 25, December 1, 1941. Hearings. GPO, Washington, DC.

United States. Office of Naval Intelligence (1946b) Ex. 149, Memorandum, Japanese fleet locations, December 1, 1941. Hearings. GPO, Washington, DC.

United States. Office of Naval Intelligence (1946c) Ex. 8, Naval message, no subject, November 24, 1941. Hearings. GPO, Washington, DC.

Tversky A, Kahneman D (1974) Judgment under uncertainty: Heuristics and biases. Science, 185(4157):1124-1131.

West, M (1982) Aspects of recursive Bayesian estimation. Ph.D. dissertation, Sch. Mathetmatical Sci., University of Nottingham, England.

Wohlstetter, R (1962) Pearl Harbor: warning and decision. Stanford University Press, Stanford, Calif.

Zlotnick J (1972) Bayes' theorem for intelligence analysis. Studies in Intelligence, 16(2): 43–52.